\documentclass[sigconf]{acmart}
\acmConference[Preprint]{Preprint}{November 2026}{Author version}
\acmBooktitle{Preprint}
\acmYear{2026}
\copyrightyear{2026}
\acmDOI{}
\acmISBN{}
\makeatletter
\patchcmd{\@mkbibcitation}{In \textit{\@acmBooktitle}}{\textit{\@acmBooktitle}}{}{\ClassError{acmart}{Unable to format the preprint citation}{}}
\makeatother
\usepackage{microtype}
\usepackage{graphicx}
\usepackage{booktabs,array,colortbl,tabularx}
\usepackage{float}
\usepackage{placeins}
\newcommand{\MainRLLmeAcc}{36.23}
\newcommand{\MainRLPmAcc}{32.79}
\newcommand{\MainDistilledLmeAcc}{73.11}
\newcommand{\MainDistilledPmAcc}{40.31}
\newcommand{\MainAdaptedSFTPmAcc}{48.86}

\newcommand{\DSMemoryTrainEpisodes}{445}
\newcommand{\DSMemoryTestEpisodes}{150}
\newcommand{\DSRagTrainEpisodes}{5,700}
\newcommand{\DSRetrievalTestQuestions}{400}

\newcommand{\AblationTopThreeAccuracy}{59.22}
\newcommand{\AblationTopTwelveAccuracy}{65.16}
\newcommand{\AblationSequentialAccuracy}{75.17}
\newcommand{\AblationSequentialSearches}{3.35}
\newcommand{\StandaloneMemoryDistilledAccuracy}{80.58}
\newcommand{\StandaloneMemoryMemAlphaAccuracy}{74.79}
\newcommand{\StandaloneRetrievalDistilledAccuracy}{37.75}
\newcommand{\StandaloneRetrievalSearchRoneAccuracy}{35.75}
\newcommand{\StandaloneRetrievalCoRAGAccuracy}{36.25}

\title{Composing Independently Distilled Memory and Retrieval Policies for Long-Term Agents}
\author{Grigorii Davydenko,
    Artem Sakhno,
    Igor Pavlov,
    Julia Belikova,
    Andrey Savchenko,
    Maksim Makarenko
}

\begin{document}

\begin{abstract}
Long-lived Web assistants require effective memory and retrieval mechanisms to access information from past user interactions. While these capabilities can be learned through specialized policies, it remains unclear whether specialists trained independently in different environments can be effectively composed without joint optimization. This paper investigates compositional transfer through policy distillation. Memory-writing and sequential retrieval policies are independently distilled from frontier-model
 action trajectories into 3B models and combined with a fixed answer generator without downstream training. On a LongMemEval-based benchmark, the distilled composition achieves 73.1\% accuracy, compared with 36.2\% for a composition of existing RL-trained specialists, and also outperforms the evaluated RL-trained compositions on PersonaMem-v2. Component ablations reveal that improvements in individual policies depend on their composition, highlighting the importance of cross-specialist compatibility. Further task-specific adaptation improves accuracy to 75.2\% on LongMemEval through RL and 48.9\% on PersonaMem-v2 through supervised distillation. These findings demonstrate the effectiveness of independent specialist distillation for modular multi-agent systems. Source code and datasets of teacher-generated memory-editing and retrieval trajectories are released to facilitate reproducibility and further research.
\end{abstract}

\maketitle

\section{Introduction}
\label{sec:introduction}

Web assistants increasingly operate across multiple sessions, searching, browsing, and acting on behalf of users~\cite{joko-2024-personal,zhang-2026-information}. Providing personalized assistance requires access to information unavailable on the open Web, including users' preferences, past decisions, and corrections. Long-term conversational memory therefore becomes an essential part of the retrieval infrastructure: agents must decide what information to retain, how to retrieve relevant evidence from accumulated interactions, and when sufficient evidence has been collected to answer a query.

\begin{figure}[t]
    \centering
    \includegraphics[width=\linewidth, height=6.8cm]{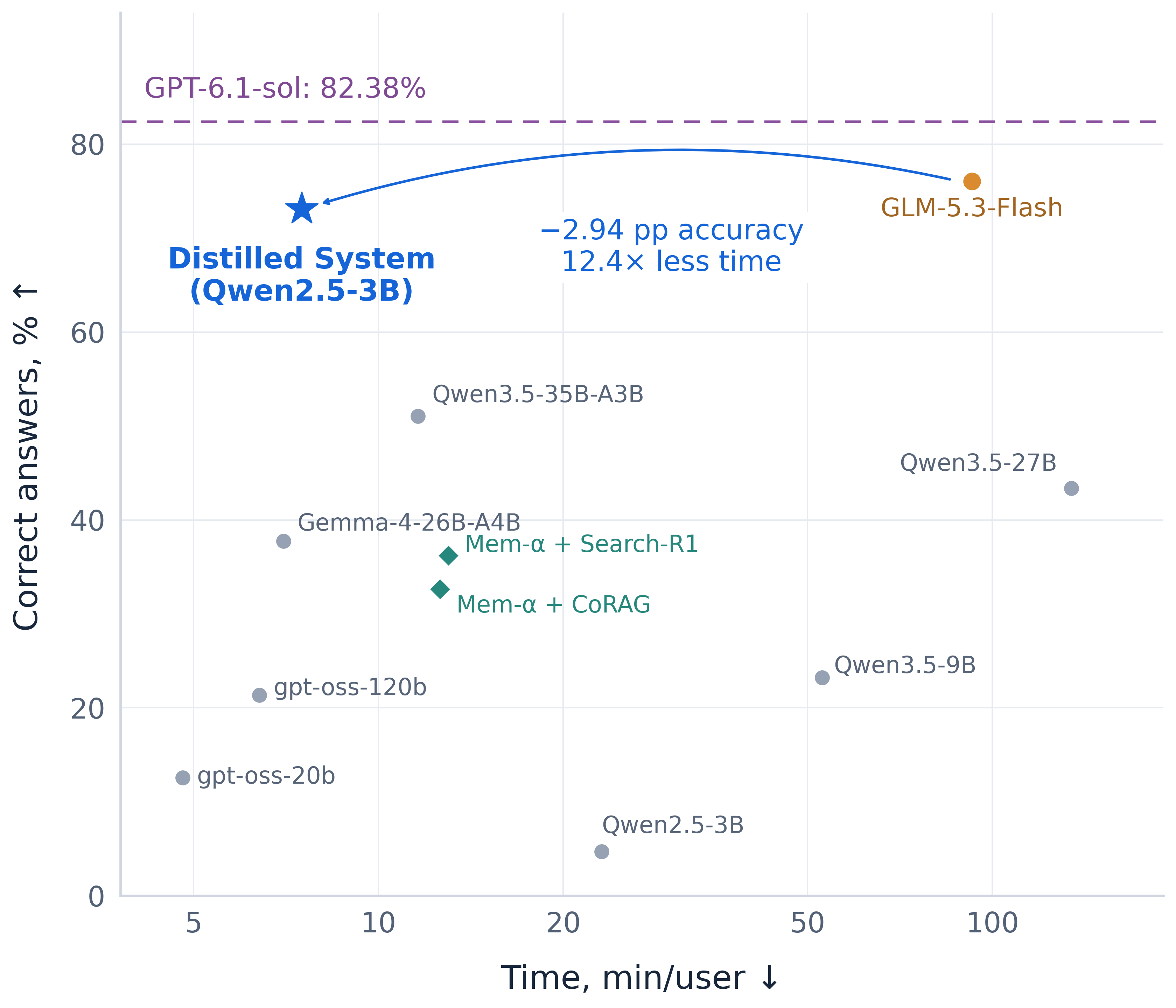}
    \caption{
    Accuracy--time trade-off on LongMemEval with a fixed GPT-6.1-sol reader. Without downstream adaptation, independently distilled specialists outperform the evaluated RL-trained compositions and approach GLM-5.3-Flash accuracy with 12.4$\times$ less processing time. The dashed line marks GPT-6.1-sol accuracy.
}
    \label{fig:accuracy-time}
    \vspace{-0.5cm}
\end{figure}

\begin{figure*}[t]
\centering
\includegraphics[width=\textwidth, height=6.5cm]{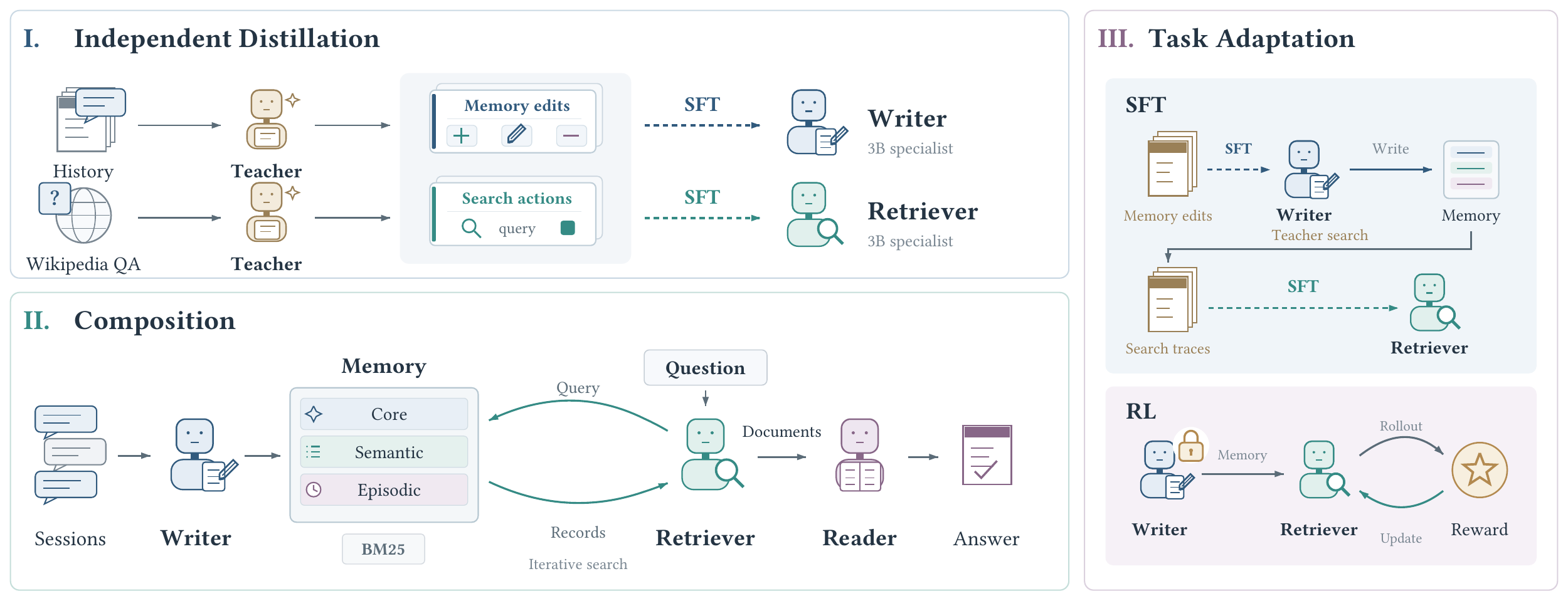}
\caption{Overview of specialist training, composition, and task-specific adaptation.
(I) Independent distillation: teacher memory edits and Wikipedia search
trajectories supervise separate writer and retriever specialists.
(II) Composition: the writer builds persistent memory from conversational
sessions; the retriever searches it iteratively, and the reader answers
from retrieved records.
(III) Task adaptation: SFT adapts the writer first, then the retriever
on teacher search traces over the resulting memory.
RL updates only the retriever, keeping the writer and memory fixed.
The reader remains fixed in both settings.}
\label{fig:workflow}
\vspace{-0.3cm}
\end{figure*}

A modular approach decomposes these responsibilities into specialized policies: a memory writer determines what information persists~\citep{wang2025memalpha}, a retriever iteratively searches memory~\citep{jin2025searchr1}, and a reader generates answers from retrieved evidence. Prior work has explored reinforcement learning for memory construction~\citep{wang2025memalpha}, sequential retrieval~\citep{jin2025searchr1,wang2025corag}, and unified memory and answering architectures~\citep{lu2023memochat}. In parallel, multi-agent research has investigated specialist training, cooperation, and knowledge distillation~\citep{bijoy2025prost,park2025maporl,lee2026mapcoderlite}, including approaches combining distillation and reinforcement learning within individual policies or predefined workflows~\citep{li2025chainofagents,wang2026lamo,yu2026chipmate}.

However, an important question remains: \emph{can specialist policies trained independently in different environments be effectively composed without joint training?} Independent training enables specialized capabilities to be developed and reused separately, potentially reducing the need for costly downstream optimization. Yet successful composition is not guaranteed. In long-term memory systems, the retriever operates over information selected and structured by the memory writer, making the effectiveness of each specialist dependent on its interaction with other components.

To investigate this question, memory-writing and sequential retrieval policies are independently distilled into 3B models using frontier-model action trajectories collected in separate environments. The resulting specialists are composed with a fixed answer generator without downstream training. Controlled component replacements examine their compositional transfer, while task-specific supervised distillation and retrieval-policy reinforcement learning are evaluated as subsequent adaptation strategies.

Experiments on LongMemEval and PersonaMem-v2 demonstrate the effectiveness of independent specialist distillation. Without downstream adaptation, the distilled composition achieves 73.11\% accuracy on LongMemEval, compared with 36.23\% for the composition of existing RL-trained specialists, and 40.31\% on PersonaMem-v2, outperforming the evaluated RL-trained compositions. Component ablations reveal that the benefits of individual policies depend on their composition, highlighting the importance of cross-specialist compatibility. Further adaptation improves accuracy to 75.17\% on LongMemEval through RL and 48.86\% on PersonaMem-v2 through supervised distillation. These findings establish independent specialist distillation as a promising approach to constructing modular multi-agent systems. Source code and datasets of teacher-generated memory-editing and sequential retrieval trajectories are released to support reproducibility and further research.

\section{Method}
\label{sec:method}

\subsection{Specialist training and composition}
\label{sec:distillation}
\label{sec:workflow}

Memory-writing and retrieval policies are distilled independently from frontier-model action trajectories collected in separate environments (Figure~\ref{fig:workflow}, I).

\noindent\textbf{Memory writing.}
The teacher processes source chunks chronologically, observing each chunk and the history of previous memory actions before generating memory-edit operations. The original Mem-$\alpha$ framework~\citep{wang2025memalpha} is retained, including its core, semantic, and episodic memory types, editing operations, and prompts.

\noindent\textbf{Sequential retrieval.}
The retrieval policy follows the Search-R1 interface~\citep{jin2025searchr1}, with the prompt adapted to require reasoning at each search step. The teacher interacts with a Wikipedia search environment, observing the question, previous queries, and retrieved documents before issuing another search query or terminating retrieval.

Teacher trajectories are converted into observation--response pairs, and two separate copies of Qwen2.5-3B-Instruct are fine-tuned to imitate the corresponding actions. Both specialists preserve their original action interfaces and are trained independently.

\begin{figure}[t]
    \centering
    \includegraphics[width=\linewidth, height=4cm]{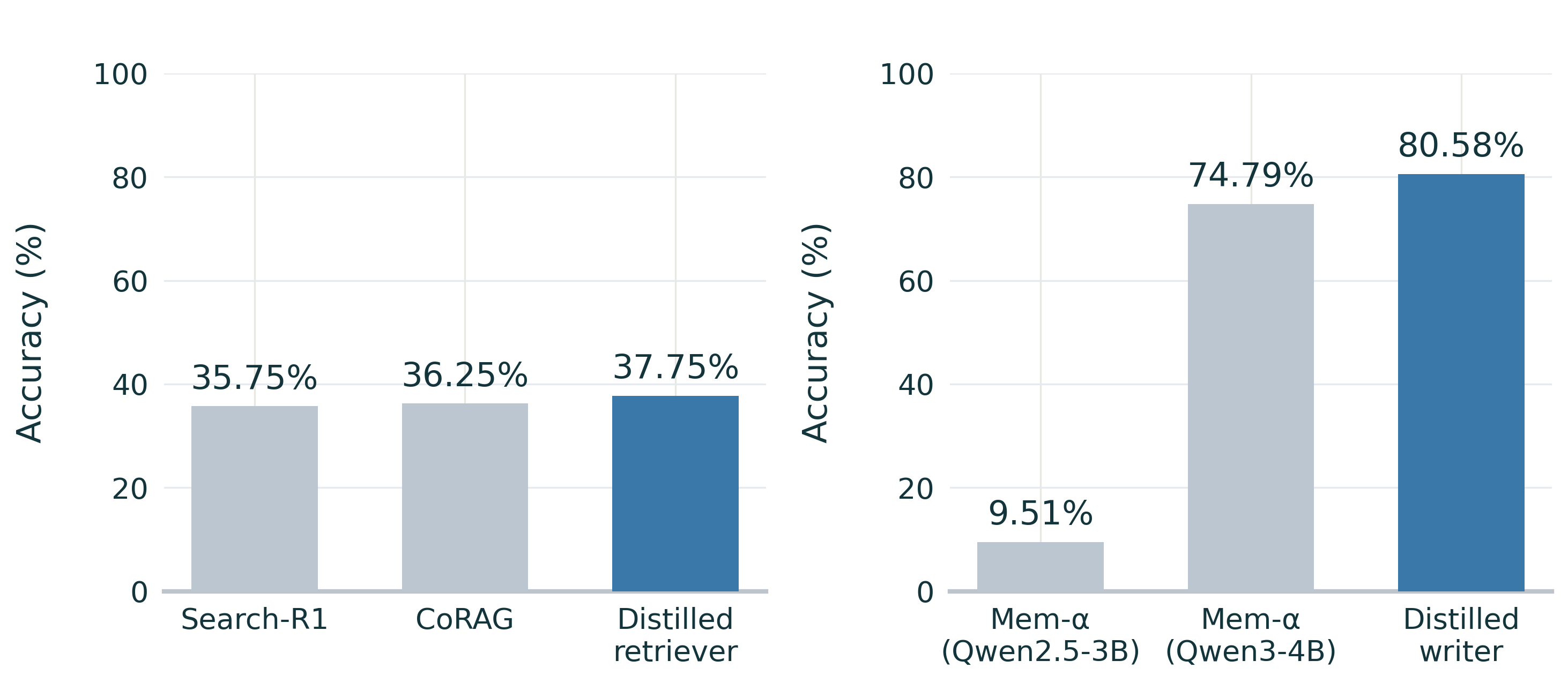}
    \caption{
        Standalone accuracy of retrieval (left) and memory (right)
        specialists before composition.
    }
    \label{fig:standalone-to-multiagent}
    \vspace{-0.3cm}
\end{figure}

\definecolor{Teal}{HTML}{315A7D}
\definecolor{Blue}{HTML}{374151}
\definecolor{Violet}{HTML}{374151}
\definecolor{PaleTeal}{HTML}{EAF0F7}
\definecolor{Muted}{HTML}{576773}

\begin{table*}[t]
\centering\small
\setlength{\tabcolsep}{3pt}
\renewcommand{\arraystretch}{1.0}
\begin{tabularx}{\linewidth}{@{}>{\raggedright\arraybackslash}p{.115\linewidth}>{\raggedright\arraybackslash}p{.115\linewidth}>{\hsize=1.0\hsize\linewidth=\hsize\centering\arraybackslash}X>{\hsize=1.00\hsize\linewidth=\hsize\centering\arraybackslash}X>{\hsize=1.00\hsize\linewidth=\hsize\centering\arraybackslash}X>{\hsize=1.00\hsize\linewidth=\hsize\centering\arraybackslash}X>{\hsize=1.00\hsize\linewidth=\hsize\centering\arraybackslash}X>{\hsize=1.00\hsize\linewidth=\hsize\centering\arraybackslash}X>{\hsize=1.00\hsize\linewidth=\hsize\centering\arraybackslash}X>{\hsize=1.00\hsize\linewidth=\hsize\centering\arraybackslash}X@{}}
\toprule
 &  & \multicolumn{4}{c}{\color{Blue}\bfseries LongMemEval} & \multicolumn{4}{c}{\color{Violet}\bfseries PersonaMem-v2} \\
Memory & Retrieval & \mbox{Accuracy (\%)} & Searches & \mbox{Memory (KB)} & \mbox{Time (min)} & \mbox{Accuracy (\%)} & Searches & \mbox{Memory (KB)} & \mbox{Time (min)} \\
\midrule
\multicolumn{10}{@{}l}{\color{Muted}\footnotesize\textsc{Specialists}} \\
\addlinespace[1pt]
Qwen2.5-3B & Qwen2.5-3B & 4.69 & 2.43 & 13.1 & 23.1 & 33.02 & 1.34 & 10.9 & 7.4 \\
Mem-$\alpha$ & Search-R1 & 36.23 & 1.40 & 175.4 & 13.0 & 32.79 & 0.86 & 307.4 & 42.6 \\
Mem-$\alpha$ & CoRAG & 32.63 & 3.95 & 175.4 & 12.6 & 36.49 & 3.91 & 307.4 & 42.6 \\
Distilled & Distilled & 73.11 & 2.84 & 94.3 & 7.5 & 40.31 & 3.36 & 232.9 & 27.8 \\
\midrule
\multicolumn{10}{@{}l}{\color{Muted}\footnotesize\textsc{Full Systems}} \\
\multicolumn{2}{@{}l}{MemoChat} & 64.22 & 1.00 & 526.7 & 2.5 & 45.33 & 1.00 & 564.0 & 4.9 \\
\multicolumn{2}{@{}l}{Directly Adapted (SFT)} & 54.08 & 1.53 & 93.8 & 20.6 & \underline{47.66} & 2.85 & 131.7 & 10.3 \\
\multicolumn{2}{@{}l}{Distilled \& Adapted (RL)} & \textbf{75.17} & 3.35 & 94.3 & 7.7 & 45.49 & 3.82 & 232.9 & 27.8 \\
\rowcolor{PaleTeal}
\multicolumn{2}{@{}l}{\textcolor{Teal}{\textbf{Distilled \& Adapted (SFT)}}} & \underline{74.68} & 1.84 & 85.1 & 8.6 & \textbf{48.86} & 3.22 & 128.0 & 9.1 \\
\midrule
\multicolumn{10}{@{}l}{\color{Muted}\footnotesize\textsc{Frontiers}} \\
\addlinespace[1pt]
GLM-5.3-Flash & GLM-5.3-Flash & 76.05 & 1.30 & 97.0 & 92.7 & 54.95 & 2.26 & 135.6 & 158.3 \\
GPT-6.1-sol & GPT-6.1-sol & 82.38 & 1.21 & 96.7 & 29.3 & 54.78 & 2.71 & 175.6 & 45.3 \\
\bottomrule\end{tabularx}
\caption{Dialogue performance on LongMemEval and PersonaMem-v2. All systems use a GPT-6.1-sol reader. Searches is the mean number of search calls per question, averaged over users. Memory is the mean stored text per user in KB. Time is the mean total time per user for memory construction and retrieval, in minutes, excluding the reader. Local runs use one A100, or eight for GLM. Bold and underlined values indicate the best and second-best accuracy among non-frontier systems.}
\label{tab:composition}
\end{table*}

\noindent\textbf{Composition.}
The distilled specialists are combined into a dialogue system without downstream training (Figure~\ref{fig:workflow}, II). The memory writer processes conversational sessions to construct a persistent memory store indexed by BM25. Given a question, the retriever iteratively searches this store until it decides to stop or exhausts the search budget. A fixed reader then generates an answer conditioned on the question and retrieved records.

\subsection{Task-specific adaptation}
\label{sec:task-adaptation}

Two adaptation strategies are investigated while keeping the reader fixed (Figure~\ref{fig:workflow}, III).

\noindent\textbf{Supervised distillation.}
Additional teacher trajectories are collected on the downstream dialogue tasks. The memory writer is first adapted through supervised fine-tuning, followed by the retriever using teacher search trajectories over the resulting memory stores.

\noindent\textbf{Reinforcement learning.}
Only the retriever is adapted, while the memory writer and its constructed memory stores remain fixed. Starting from the distilled checkpoint, the retriever is optimized using Group Relative Policy Optimization (GRPO) with eight trajectories per question. The reward is final answer correctness.

\section{Experiments}
\label{sec:experiments}

\subsection{Training and standalone evaluation}
\label{sec:training-and-evaluation}

Both specialists are initialized from Qwen2.5-3B-Instruct and distilled from GPT-6.1-sol trajectories with medium reasoning effort. Memory training uses the Mem-$\alpha$ collection \citep{wang2025memalpha} with its original split of \DSMemoryTrainEpisodes/\DSMemoryTestEpisodes\ train/test episodes. Retrieval uses Search-R1 data \citep{jin2025searchr1}: \DSRagTrainEpisodes\ training questions and \DSRetrievalTestQuestions\ held-out questions, balanced across Natural Questions, HotpotQA, 2WikiMultiHopQA, and MuSiQue.

We first evaluate the specialists separately, before composition
(Figure~\ref{fig:standalone-to-multiagent}).
The distilled retriever achieves
\StandaloneRetrievalDistilledAccuracy\% accuracy, compared with
\StandaloneRetrievalSearchRoneAccuracy\% for Search-R1 and
\StandaloneRetrievalCoRAGAccuracy\% for CoRAG (both based on Qwen2.5-3B-Instruct).
The distilled memory writer achieves
\StandaloneMemoryDistilledAccuracy\%, compared with
\StandaloneMemoryMemAlphaAccuracy\% for the released Mem-$\alpha$~\citep{wang2025memalpha}
checkpoint based on Qwen3-4B.
For a controlled comparison, we additionally trained Mem-$\alpha$
using the same Qwen2.5-3B-Instruct backbone as the distilled writer.
This configuration achieved 9.51\% accuracy on the original
Mem-$\alpha$ test set, compared with 8.34\% for the untrained
model. Since this configuration did not yield a competitive
baseline, subsequent comparisons use the officially released
Mem-$\alpha$ checkpoint based on Qwen3-4B.
GPT-6-luna with maximum reasoning effort serves as the judge
for both standalone evaluations.

\subsection{Evaluating specialist composition}
\label{sec:experimental-setup}

We evaluate conversational QA on 147 LongMemEval S questions
covering three users and 370 official PersonaMem-v2
multiple-choice questions covering 20 users.
Training and test sets have no overlapping
sessions, including those repeated across users.

Table~\ref{tab:composition} compares three groups:
(1) specialist compositions without downstream adaptation:
Qwen2.5-3B-Instruct in both roles, the released Mem-$\alpha$
writer paired with Search-R1 \citep{jin2025searchr1} or CoRAG,
and our independently distilled pair;
(2) full systems: MemoChat, direct SFT of Qwen2.5-3B-Instruct
without prior distillation, and our distilled pair with
retriever-only RL or SFT of both specialists;
and (3) frontier pipelines using GLM-5.3-Flash or GPT-6.1-sol
for both memory writing and retrieval without task-specific
fine-tuning.
MemoChat (initialized from Qwen2.5-3B-Instruct) is fine-tuned separately for each benchmark
using GPT-6.1-sol examples.

All systems share a GPT-6.1-sol reader with medium reasoning
effort, supplied with all retrieved records; this replaces
MemoChat's answering stage.

Released specialists retain their original action interfaces.
Sequential retrieval allows up to four searches per question,
returning up to three records each.
We report user-averaged accuracy, search counts, memory size,
and runtime.
LongMemEval answers are judged by GPT-6-luna with max reasoning
effort and PersonaMem-v2 uses its official answer key.

Without downstream adaptation, the distilled pair achieves
\MainDistilledLmeAcc\% accuracy on LongMemEval, compared with
\MainRLLmeAcc\% for Mem-$\alpha$ + Search-R1. On PersonaMem-v2,
the corresponding accuracies are \MainDistilledPmAcc\% and
\MainRLPmAcc\%. The distilled pair also outperforms the
MemoChat-based pipeline on LongMemEval. Among non-frontier
systems, Distilled \& Adapted (RL) achieves the highest LongMemEval
accuracy, while Distilled \& Adapted (SFT) achieves the highest
PersonaMem-v2 accuracy at \MainAdaptedSFTPmAcc\%.
The distilled specialists therefore form an effective composition
even without downstream adaptation.

\subsection{Additional experiments}
\label{sec:sequential-retrieval}

\definecolor{Teal}{HTML}{315A7D}\definecolor{PaleTeal}{HTML}{EAF0F7}
\begin{table}[t]\centering\small
\begin{tabularx}{\linewidth}{@{}l*{2}{>{\centering\arraybackslash}X}@{}}
\toprule
Configuration & Accuracy (\%) & Searches \\
\midrule
BM25 top-3 & 59.22 & 1.00 \\
BM25 top-12 & \underline{65.16} & 1.00 \\
Qwen2.5-3B & 64.92 & 2.12 \\
\rowcolor{PaleTeal}
\textcolor{Teal}{\textbf{Distilled \& Adapted (RL)}} & \textbf{75.17} & 3.35 \\
\bottomrule\end{tabularx}
\caption{Sequential retrieval versus single-query BM25 on LongMemEval. All configurations use the same memory and reader.}
\label{tab:sequential-retrieval}
\vspace{-0.3cm}
\end{table}

\noindent\textbf{Sequential retrieval.}
Table~\ref{tab:sequential-retrieval} compares the RL-adapted retriever with single-query BM25 and an unadapted Qwen2.5-3B retriever over the same memory stores constructed by the distilled writer. While base Qwen2.5-3B performs iterative search (2.12 searches on average), it reaches only 64.92\% accuracy---comparable to BM25 top-12 (\AblationTopTwelveAccuracy\%) and above BM25 top-3 (\AblationTopThreeAccuracy\%). By contrast, the RL-adapted retriever achieves \AblationSequentialAccuracy\% using \AblationSequentialSearches\ searches (+10.25\,pp over base Qwen2.5-3B), confirming that gains stem from learned query refinement rather than multiple search turns alone.

\begin{figure}[H]
\centering
\includegraphics[width=\linewidth]{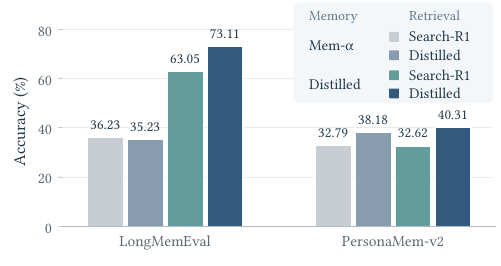}
\caption{Component impact ablation. Single-specialist replacements show uneven transfer, whereas pairing both distilled specialists maximizes accuracy on both benchmarks.}
\label{fig:component-ablation}
\vspace{-0.3cm}
\end{figure}

\noindent\textbf{Component impact.}
Figure~\ref{fig:component-ablation} isolates the effect of
replacing each specialist while keeping the reader fixed.
On LongMemEval, distilled retrieval reduces accuracy when
paired with Mem-$\alpha$ (36.23\% to 35.23\%) but improves
it substantially with distilled memory (63.05\% to 73.11\%).
On PersonaMem-v2, distilled retrieval improves both compositions,
whereas replacing the memory writer yields mixed results.
The fully distilled pair achieves the highest accuracy among
the four unadapted combinations on both benchmarks,
highlighting the importance of cross-specialist compatibility.

\FloatBarrier

\begin{figure}[t]
    \centering
    \includegraphics[width=\linewidth]{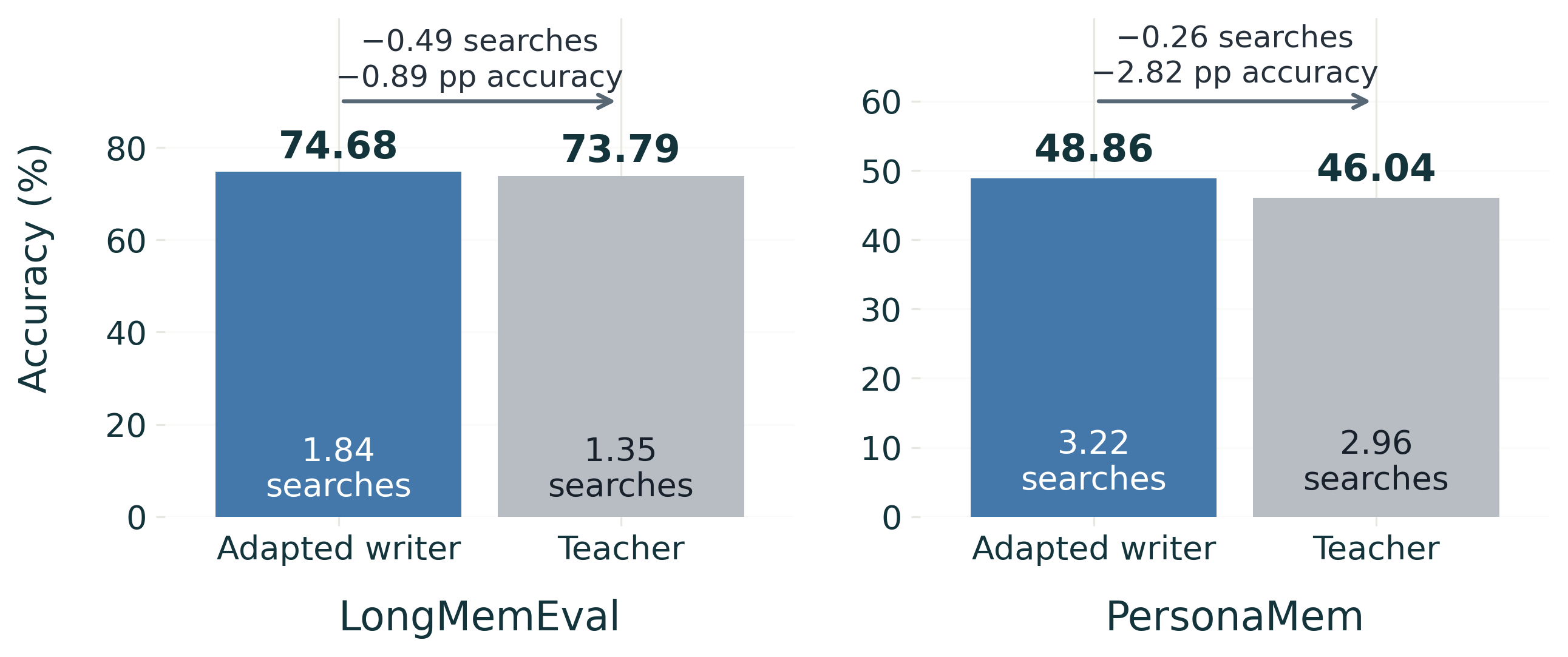}
    \caption{
        Effect of the memory source used for retriever SFT.
        Both variants use the adapted student writer at evaluation.
    }
    \label{fig:sft-memory-source}
    \vspace{-0.3cm}
\end{figure}

\noindent\textbf{Memory source for retriever SFT adaptation.}
The default SFT procedure first adapts the memory writer,
then trains the retriever on teacher trajectories over the
adapted writer's stores. Using teacher-written stores instead
reduces accuracy by 0.89 percentage points on LongMemEval
and 2.82 percentage points on PersonaMem-v2
(Figure~\ref{fig:sft-memory-source}).
This suggests that matching the memory distributions
encountered during retriever training and inference
is important for effective adaptation.

\section{Conclusion}

This work demonstrates that independently distilled memory-writing and retrieval policies can be effectively composed into a multi-agent system without joint training. The distilled specialists outperform the evaluated compositions of existing RL-trained policies on LongMemEval and PersonaMem-v2. Component ablations reveal that downstream performance depends on cross-specialist compatibility, rather than individual policy quality alone. Further adaptation through supervised distillation and reinforcement learning provides additional improvements, although a gap to frontier models remains, particularly in personalization. These findings highlight independent specialist distillation as a promising approach to building modular multi-agent systems and motivate further research into compositional generalization.

\end{document}